\pdfoutput=1
\documentclass[11pt]{article}
\usepackage{tabularx}
\usepackage{graphicx}
\usepackage{array}
\usepackage{booktabs}
\usepackage{longtable}
\usepackage{ragged2e}
\usepackage{tikz}
\usepackage{hyperref}
\usepackage{comment}

\usetikzlibrary{shapes, arrows.meta, positioning, backgrounds, fit} 
\usepackage{adjustbox}
\usepackage{lipsum}
\usepackage{natbib}
\usepackage{xcolor}

\usepackage[final]{acl}

\usepackage{times}
\usepackage{latexsym}

\usepackage[T1]{fontenc}

\usepackage[utf8]{inputenc}

\usepackage{microtype}

\usepackage{inconsolata}

\usepackage{graphicx}

\usepackage{pifont}
\makeatletter
\renewcommand{\@fnsymbol}[1]{%
  \ifcase#1\or
    \ding{41}
  \else
    \@arabic{#1}
  \fi
}
\makeatother

\title{Trustworthy Medical Question Answering: An Evaluation-Centric Survey}

\author{
  \textbf{Yinuo Wang\textsuperscript{1}} \space\space\space
  \textbf{Baiyang Wang\textsuperscript{1}} \space\space\space
  \textbf{Robert E. Mercer\textsuperscript{2}} \space\space\space
  \textbf{Frank Rudzicz\textsuperscript{3,4,5}} \space\space\space
  \\
  \textbf{Sudipta Singha Roy\textsuperscript{2}} \space\space\space
  \textbf{Pengjie Ren\textsuperscript{1}} \space\space\space
  \textbf{Zhumin Chen\textsuperscript{1}} \and
  \textbf{Xindi Wang \textsuperscript{1}}\thanks{Corresponding author} \\
  \textsuperscript{1} Shandong University, China \space\space\space
  \textsuperscript{2} University of Western Ontario, Canada \\
  \textsuperscript{3} Dalhousie University, Canada \space\space\space
  \textsuperscript{4} University of Toronto, Canada \\
  \textsuperscript{5} Vector Institute for Artifcial Intelligence, Canada \\
  \texttt{\{202420871, 202320482\}@mail.sdu.edu.cn}\\
  \texttt{ mercer@csd.uwo.ca, frank@dal.ca, ssinghar@uwo.ca}\\
  \texttt{\{renpengjie, chenzhumin, xindi.wang\}@sdu.edu.cn}
}

\begin{document}

\maketitle
\begin{abstract}
%
Trustworthiness in healthcare question-answering (QA) systems is important for ensuring patient safety, clinical effectiveness, and user confidence. As large language models (LLMs) become increasingly integrated into medical settings, the reliability of their responses directly influences clinical decision-making and patient outcomes. However, achieving comprehensive trustworthiness in medical QA poses significant challenges due to the inherent complexity of healthcare data, the critical nature of clinical scenarios, and the multifaceted dimensions of trustworthy AI. In this survey, we systematically examine six key dimensions of trustworthiness in medical QA, i.e., Factuality, Robustness, Fairness, Safety, Explainability, and Calibration. 
We review how each dimension is evaluated in existing LLM‐based medical QA systems. We compile and compare major benchmarks designed to assess these dimensions and analyze evaluation‐guided techniques that drive model improvements, such as retrieval‐augmented grounding, adversarial fine‐tuning, and safety alignment. Finally, we identify open challenges, such as scalable expert evaluation, integrated multi‐dimensional metrics, and real‐world deployment studies, and propose future research directions to advance the safe, reliable, and transparent deployment of LLM‐powered medical QA.

%
\end{abstract}

\section{Introduction}

\begin{figure*}[htbp] 
  \centering
  \includegraphics[width=\textwidth]{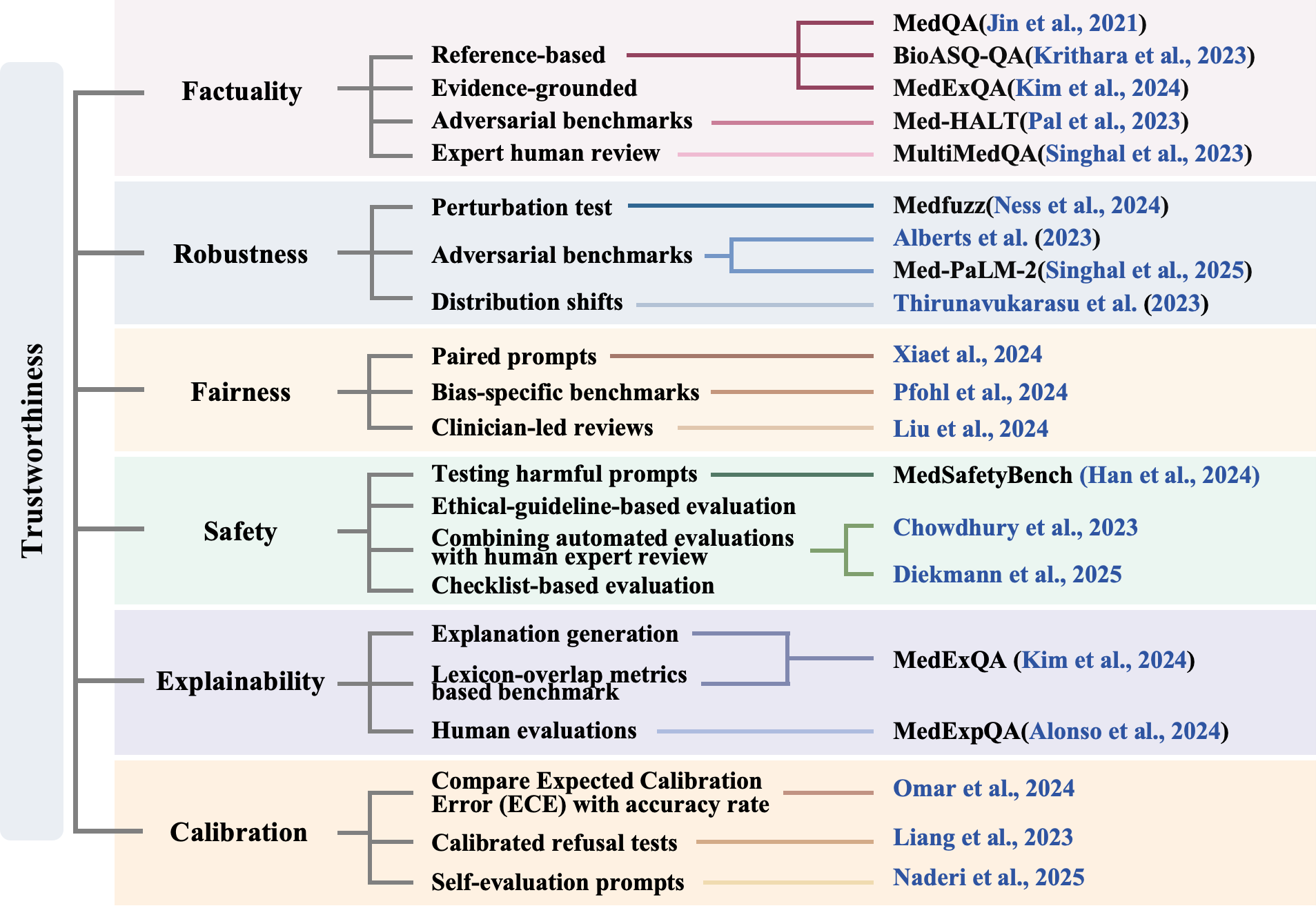} 
  \caption{Taxonomy of Evaluation Dimensions of Trustworthiness. The taxonomy includes six core dimensions, each with corresponding assessment methods. For each method, representative benchmarks are provided. }
  \label{fig:1}
\end{figure*}

Large language models (LLMs) have significantly advanced the field of question-answering (QA)  \cite{MINT,Eva}, enabling remarkable capabilities in generating fluent and coherent responses across a wide range of domains. In healthcare, specialized variants such as Med‐PaLM~\cite{MultiMedQA} and ChatDoctor~\cite{Chatdoctor} have even matched or exceeded human performance on professional exams —Med‐PaLM achieved a passing score of 67.6\% on USMLE-style MedQA questions and Med-PaLM 2 reached 86.5\% accuracy—
and have demonstrated superior consumer‐health assistance in user studies~\cite{zhongjing,lhmr}. Yet, when deployed in clinical settings, these models continue to exhibit critical trust failures: hallucinated medical facts, unjustified overconfidence, and occasional biased or unsafe recommendations~\cite{1}. Such errors can directly endanger patient safety, lead to misdiagnoses, or exacerbate healthcare disparities, underscoring that trustworthiness in medical QA is not optional but essential. 

Although recent surveys have mapped broad trust dimensions—truthfulness, safety, robustness, fairness, and explainability—for LLMs in healthcare, work focused specifically on open‐domain medical QA remains fragmented~\cite{2,3,4}. Existing reviews typically catalogue each dimension in isolation, without clearly linking evaluation findings to concrete model improvements. In practice, a single evaluation signal often indicates multiple risks, yet this interplay is seldom analyzed or leveraged to guide system development holistically.
To bridge this gap, 
we adopt an evaluation-driven framework tailored specifically for medical QA. We first define six core dimensions—Factuality 
, Robustness, Fairness, Safety, Explainability, and Calibration—and consolidate the primary evaluation methods for each into a unified taxonomy, shown in Figure \ref{fig:1}. We then demonstrate how evaluation insights have directly  inspired targeted optimizations. Building on this, we review the 
benchmarks and tools, comparing their methodological trade offs. Finally, we examine open challenges and propose future research directions. By weaving together evaluation, optimization, and benchmarking, our survey provides a clear roadmap for leveraging trustworthiness assessments as catalysts for building safer, more reliable, and equitable LLM-powered medical QA systems.



\section{Evaluation Dimensions of Trustworthiness}

Trustworthiness in medical QA is inherently multidimensional, encompassing various interconnected evaluation criteria. In this section, we 
define six core dimensions for assessing trustworthiness specifically within medical QA contexts. 

\subsection{Factuality} 


Factuality evaluates whether a medical QA system’s responses are both correct and verifiable against established clinical knowledge, inherently encompassing the detection of hallucinations—plausible-sounding but unsupported or incorrect statements \cite{factualitysurvey,hallucinationsurvey}. Even minor factual errors in healthcare can compromise patient safety, so rigorous evaluation is indispensable.

Assessment often begins with reference-based measures. For structured tasks such as USMLE-style multiple-choice questions~\cite{MedQA}, simple accuracy suffices. For open-ended responses, metrics like Exact Match or token-overlap F1 are calculated against curated reference answers~\cite{BioASQ-QA}. To accommodate valid variability in medical phrasing, benchmarks frequently allow lenient scoring or use multiple expert-generated references, as in MedExQA’s ensemble of clinician explanations~\cite{MedExQA}. Evidence-grounded checks then verify that each factual claim can be traced back to authoritative sources—peer-reviewed articles, clinical guidelines, or trusted medical databases—flagging unsupported content as potential hallucinations. Adversarial benchmarks like Med-HALT~\cite{Med-HALT} and targeted “false-confidence” probes stress-test models with challenging prompts designed to induce fabrications, thereby quantifying a model’s propensity to hallucinate under duress. 
Because factuality in medicine can sometimes be a ``grey area'', especially when clinical guidelines evolve or expert consensus varies, automated metrics alone may not suffice~\cite{landsheer2018clinical}. In such cases, expert human review remains the gold standard: clinicians apply structured rubrics (for example, the Med-PaLM evaluation framework~\cite{MultiMedQA}) to rate answers on accuracy, completeness, and consistency with medical consensus. This catches subtle inaccuracies and context-specific errors that automated metrics may miss.

These approaches form a comprehensive framework for measuring factual accuracy and hallucination in medical QA. The insights they provide directly inform mitigation techniques such as retrieval-augmented grounding to anchor responses in live literature, post-hoc fact-correction modules to revise unsupported claims, adversarial fine-tuning to harden models against deceptive inputs, and iterative self-reflection loops that internally check for consistency—collectively advancing the safety and reliability of medical QA systems.

\subsection{Robustness}
\label{sec:Robustness}

Robustness refers to the system’s ability to maintain performance under varied 
inputs in medical QA. A robust model should handle paraphrased questions, out-of-distribution queries, or adversarial inputs without significant degradation in answer quality~\cite{robustnesssurvey,robusurvey}.

One way to measure robustness is by perturbing real queries—rephrasing symptom descriptions, introducing spelling mistakes, or inserting extraneous clauses—and then checking whether the model’s output remains correct. For example, \citet{Medfuzz} introduced MedFuzz, a method designed to systematically perturbed medical questions to test whether models depend on superficial linguistic patterns. Their findings indicate that even subtle variations in phrasing can disrupt a model’s reasoning process, thus exposing inherent brittleness. Another key aspect is adversarial robustness, which entails ensuring that models are resilient to intentionally deceptive or challenging inputs. In medical QA, adversarial scenarios may involve misleading cues that integrate multiple complex concepts. \citet{alberts2023large} emphasized that adversarial testing in medical QA must account for the inherent complexity of the domain, noting that even slight modifications in phrasing 
can significantly alter clinical interpretations. Evaluations may incorporate challenge sets comprising known difficult cases, such as rare conditions or overlapping symptoms, to assess model performance comprehensively. For instance, the Med-PaLM-2 study specifically included a set of adversarial questions designed to probe the limitations of LLMs, which can be used to conduct targeted evaluations to identify cases that intentionally elicit confusion or highlight model vulnerabilities~\cite{MedPaLM2}. Robustness can also be characterized by resilience to distributional shifts, referring to a model’s ability to maintain performance when encountering inputs that differ substantially from its training data. For example, a model trained primarily on formal medical texts may struggle with questions phrased in layperson language. Consequently, evaluators often test models using cross-style or cross-population datasets, including questions derived from different demographic groups or varied linguistic styles. Sustained model performance under these conditions indicates robustness against such distributional shifts. Quantitatively, robustness can be measured by the performance drop observed when transitioning from clean to perturbed datasets; a minimal decline reflects higher robustness. Additionally, variance-based measures are employed; for instance, \citet{thirunavukarasu2023large} proposed evaluating the variance in model outputs across semantically equivalent inputs as an indicator of robustness.

Comprehensive robustness evaluation guides improvements like adversarial fine-tuning, data augmentation with diverse linguistic styles, and multi-domain training, ultimately yielding more stable and trustworthy medical QA systems.

\subsection{Fairness}
Fairness in medical QA assesses whether a system's performance is equitable across diverse user groups and contexts, avoiding biased or stereotypical responses. In medicine, fairness concerns 
involve patient demographics
, health conditions, or socioeconomic factors~\cite{Fairbiassv}. An unfair system may provide inconsistent answers based on demographic attributes or reflect biases from training data~\cite{Fairsurvey}.
Crucially, fairness evaluation 
must distinguish harmful social biases 
from medically justified, evidence-based demographic differences
~\cite{Acausal}. 
For instance, the higher prevalence of sickle cell anemia in individuals of African descent is a clinically relevant demographic pattern that should be preserved, not a form of algorithmic bias to be mitigated.
Evaluating fairness is challenging because biases can be subtle or implicit. One effective technique uses paired prompts that differ only in a demographic detail—such as ``What is the best treatment for a male patient with symptom $X$?'' versus ``a female patient with symptom $X$?''—to detect 
discrepancies in content, confidence, or thoroughness. Empirical studies have shown medical LLMs often vary their recommendations across demographic groups, reflecting biases in their training data~\cite{xia2024cares}. Additional methods include bias-specific benchmarks (race-focused or condition-focused query sets) and clinician-led reviews where experts flag any stereotype or inequitable treatment~\cite{liu2024large}. Quantitative metrics like group-wise accuracy gaps and qualitative bias annotations help reveal fairness issues~\cite{biasmt1}. However, a major obstacle is the lack of large, bias-annotated medical QA corpora—most evaluations rely on small, hand-crafted case sets or retrospective analyses of model outputs.

To address these gaps, future work should invest in building extensive, demographically diverse fairness benchmarks and incorporate fairness-aware techniques into model training—such as data-augmentation for under-represented groups, adversarial debiasing, and fairness constraints. These combined strategies will help ensure AI-driven medical QA delivers accurate, respectful, and equitable guidance to every patient.

\subsection{Safety}

Safety evaluation assesses whether a medical QA system's responses avoid causing harm. In a medical context, unsafe answers could 
encourage harmful actions (e.g., discontinuing medication without consultation), give illegal or unethical advice, violate privacy, or otherwise contravene medical ethics~\cite{Medsafetybench}. Safety evaluations often verify that models appropriately refuse or handle unsafe requests and ensure their responses contain no harmful content
~\cite{Safesurvey,Safemed,ChLLMsafety}.

A practical method for evaluating model 
safety involves testing responses to harmful user queries, such as requests for prescription drugs without authorization or unsafe medical advice. MedSafetyBench~\cite{Medsafetybench} provides 
harmful medical prompts paired with safe responses. It 
shows that LLMs often fail safety standards and demonstrate improvements through fine-tuning. Automated evaluations using content filters or classifiers can detect overtly harmful responses, but nuanced medical contexts require human expert reviews. Experts ensure responses address medical issues safely and include essential warnings~\cite{chowdhury-etal-2023-large}. Additionally, model outputs should align with ethical guidelines, such as AMA's medical ethics principles—autonomy, non-maleficence, beneficence, and justice. Evaluations typically use checklists to assess harmfulness, encouragement of unprofessional actions, and privacy concerns. 


\subsection{Explainability} 
\label{sec:Explainability and Reasoning}

Explainability evaluates how well the system can provide reasoning or justification for its answers~\cite{explainabilitysv}. In medical QA, explanations are vital: clinicians and patients are more likely to trust an answer if they understand why the model gave it. Moreover, a correct answer without rationale may be less useful in practice than a slightly incomplete answer with a solid explanation that a clinician can follow up on. 

Explainability assessments 
involve two aspects: the presence of explanations and their quality—accuracy and clarity. Benchmarks such as MedExQA~\cite{MedExQA} explicitly require models to provide explanations, comparing them against multiple ground-truth explanations using lexical metrics (e.g., BLEU/ROUGE). However, lexical overlap alone isn't sufficient, as fluent explanations might still be incorrect or irrelevant. Thus, human evaluations are essential, with experts rating explanations for correctness, completeness, and coherence. However, a critical challenge in explainability is the distinction between an explanation's plausibility and its faithfulness. Plausibility refers to how convincing an explanation appears to humans, whereas faithfulness measures how accurately it reflects the model's true internal reasoning process \cite{Faithfulness}. Many post-hoc explanation techniques, such as saliency maps or attention weights, can produce justifications that are superficially plausible but do not faithfully capture the model’s actual decision-making mechanisms. Therefore, even when such explanations enhance user trust or perceived usability, they risk being misleading if interpreted as a faithful indicators of model reasoning. \citet{MedExpQA} included human annotation in MedExQA and demonstrated that models offering better explanation correlated with deeper understanding.

Explainability also extends to complex tasks requiring detailed reasoning, such as multi-hop questions or diagnostic case studies \cite{feng-etal-2020-scalable}. Transparent and consistent explanations indicating clear logic receive higher ratings. Evaluating explanation quality ensures that  models truly understand medical content rather than simply guessing correctly, thus enhancing trust and practical utility~\cite{reasoningsv}.

\subsection{Calibration}

Calibration in medical QA refers to how well a model's confidence aligns with the accuracy of its answers~\cite{calibration1,calibrmt2}. 
A well-calibrated model recognizes the limits of its knowledge, expressing high confidence when correct and appropriate uncertainty when potentially incorrect.
Effective calibration is critical in medicine, as overly confident yet incorrect answers pose serious risks, while excessive uncertainty limits usability.

Calibration evaluation 
involves comparing the model’s expressed confidence to its actual accuracy \cite{OnCalibration}. Metrics include comparing stated confidence levels to accuracy rates and Expected Calibration Error (ECE), which quantifies discrepancies between predicted confidence and observed accuracy; lower ECE indicates better calibration. Practically, evaluators test calibration using questions of varying difficulty. A model should confidently answer straightforward questions but express uncertainty for complex
cases. \citet{liang2023holistic} introduced calibrated refusal tests, which formalize the use of abstention by requiring models to appropriately indicate uncertainty or refuse to answer challenging questions. Another method involves self-evaluation prompts, where models assess their confidence post-response. Good calibration means models recognize and express uncertainty when their answers might be incorrect. Recent research explored integrating uncertainty quantification into LLMs to improve calibration, enhancing the correlation between confidence and correctness~\cite{1}.

Ultimately, strong calibration reduces the risk of confidently incorrect responses, enabling safer clinical use by allowing models to employ abstention methods for unsafe or low-confidence queries, or by otherwise clearly indicating when human intervention or review is necessary.

\subsection{Interplay Among Trustworthiness Dimensions}
Although we define the six dimensions as distinct evaluation axes, real-world medical QA systems exhibit important cross‐dimension interactions that can be exploited for more holistic improvements.

\textbf{Factuality and Calibration} Hallucinations almost always coincide with misplaced confidence. \citet{10.1145/3618260.3649777} show that ``hallucination'' set a statistical lower bound on calibration error in LLMs, and that techniques which reduce overconfidence also diminish hallucination rates. By training models to express uncertainty when evidence is lacking, we see both better calibration curves and fewer factual errors.

\textbf{Robustness and Factuality} Models fine‐tuned to resist adversarial or paraphrased inputs (e.g., via MedFuzz‐style perturbations) demonstrate lower hallucination rates, since they rely less on spurious patterns~\cite{asgari2025framework}. Robustness training thus directly curtails factual errors by enforcing consistency under input variations.

\textbf{Fairness and Safety} Biased medical advice (e.g., underestimating pain in certain demographics) not only undermines equity but can lead to unsafe under‐treatment. Studies of demographic bias in medical LLMs show that fairness interventions (such as adversarial debiasing) reduce both performance gaps and harmful, biased recommendations~\cite{walsh_2024}. Ensuring equitable answers therefore bolsters overall patient safety.

\textbf{Explainability and Calibration} Transparent justifications help users and downstream evaluators assess a model’s certainty. Umapathi et al. demonstrate that sample‐consistency methods—prompting the model to generate and compare multiple reasoning chains—both improve calibration and produce more faithful explanations~\cite{savage2024large}. When a model clearly cites its reasoning, confidence estimates align more closely with actual correctness.

\textbf{Calibration and Safety} Overconfident responses to high‐risk medical queries can directly endanger patients. The MedSafetyBench benchmark finds that models with tighter confidence thresholds refuse unsafe advice more reliably~\cite{Medsafetybench}. Thus, calibration improvements (e.g., via atypicality‐aware recalibration reducing ECE by 60\%) yield safer behaviour.

Understanding these synergies allows us to design multi‐axis evaluation suites—for example, safety tests stratified by confidence levels or robustness checks across demographic groups—that reveal a model’s trust profile more fully. Moreover, optimization strategies (such as retrieval‐augmentation or adversarial fine‐tuning) can be prioritized for their compound benefits across several dimensions, leading to more reliable, equitable, and safe medical QA systems.

\section{Evaluation-Guided System Improvement for Medical QA}

A core theme in recent research is using evaluation findings to guide the development of more trustworthy medical QA systems. Rather than treating evaluation as an afterthought, the idea is to create a feedback loop: identify weaknesses via evaluation and then apply targeted improvements to the model or system design. We discuss several examples where evaluation results directly informed system changes to address each dimension.

\textbf{Reducing Hallucinations via Retrieval}
If evaluation reveals frequent factual errors or hallucinations, one solution is to supply the model with reliable external knowledge. This strategy, known as retrieval-augmented generation (RAG, ~\citet{Retrieval-Augmented,RetrievalAugmented}), has become prominent for mitigating hallucinations~\cite{chu2025reducing}.
Almanac~\cite{hallumt4} uses RAG frameworks to convert clinical QA tasks into search and retrieval processes, which use LLMs for knowledge distillation from authoritative medical sources to minimize hallucination risks. 
Similarly, an approach integrating RAG with the Negative Missing Information Scoring System (NMISS) has been effectively employed in  healthcare chatbots, providing integrated solutions for hallucination detection and reduction \cite{hallumt5}. 
Additionally, CardioCanon, a cardiology-focused chatbot, leverages RAG to ensure the accuracy and reliability of cardiological responses \cite{hallumt6}. 
Evaluation can inform retrieve strategies, for instance, if analysis shows hallucinations mostly occur on questions about rare diseases, a database for rare diseases can be linked specifically for those queries.

\textbf{Robustness through Adversarial Training}
Evaluation may show a model is brittle on certain phrasings or adversarial questions. To address this, adversarial training is used. For instance, \citet{robumt1} proposed an adversarial training framework  targeting both character-level and word-level perturbations. By systematically integrating adversarial samples into training, this approach improves robustness and generalization in  biomedical NLP tasks, including medical QA. Similarly, \citet{robumt3} explored adversarial methods via prompt engineering and fine-tuning, revealing critical model vulnerabilities and showing that adversarial fine-tuning can significantly impact model weights, an observation meriting further study. A powerful example of evaluation-guided robustness improvement involves combining MedFuzz ~\cite{Medfuzz} with targeted adversarial fine-tuning. MedFuzz is an adversarial robustness evaluation framework that systematically probes medical LLMs by generating subtle, clinically plausible perturbations to benchmark questions. These perturbations that may involve rephrasings, the addition of extraneous details, or the insertion of minor factual distractors, often induce measurable drops in accuracy or consistency, thereby revealing concrete model vulnerabilities. To address these weaknesses, the evaluation results can be used to guide adversarial fine-tuning. Specifically, Perturbation-Demonstrated Weakness Sampling (PDWS) \cite{robumt2} prioritizes the most informative adversarial examples identified by MedFuzz, ensuring that fine-tuning emphasizes cases where the model is most brittle. This integration of evaluation and training reduces performance degradation under perturbations and exemplifies how systematic adversarial assessment can drive more robust model development.

\textbf{Fairness via Data and Prompt Design} Fairness evaluation in medical QA must capture both dataset-induced biases and user-centered harms. EquityMedQA introduces seven adversarial datasets and human evaluation rubrics to measure disparities across race, gender, and geography, revealing subtle inequities in LLM responses \cite{biasmt1}. Complementary studies expose model tendencies to perpetuate debunked race-based practices \cite{biasmt2} and demonstrate how cognitive biases embedded in user inputs can distort model outputs—an effect quantified by BiasMedQA through bias-laden prompts and error analysis \cite{BiasMedQA}. Together, these benchmarks highlight uneven performance across demographic groups and underscore the need for comprehensive, multi-dimensional fairness assessments. Building on these insights, developers apply evaluation-guided interventions to mitigate unfair behaviour. Data diversification techniques—such as augmenting underrepresented groups, counter-bias pairing, and re-balancing skewed corpora—have proven effective at reducing differential performance \cite{biasmt3}. Fairness regularization and constraint-based training further enforce balanced treatment across identity attributes. At inference time, prompt engineering (e.g., “Provide gender-neutral explanations for all patients”) and user-centric guidance can nudge models toward equitable outputs, with follow-up studies showing prompt designs that specifically address cognitive biases \cite{biasmt4}. Crucially, each mitigation step is validated through repeated unbiased evaluation, forming a feedback loop: evaluate on an expanding suite of bias tests, apply targeted fixes, then re-evaluate to ensure that gains in one area do not introduce new disparities. Because real-world patients may unknowingly input misleading or biased information, future work must 
integrate robustness evaluations alongside fairness to build 
trustworthy medical QA systems.

\textbf{Alignment and Fine-Tuning for Safety}
Effective safety evaluation in medical QA combines benchmark datasets and human-aligned tests to quantify harmful-response rates and categorize unsafe behaviours. For example, MedSafetyBench supplies standardized unsafe scenarios that highlight failure modes and serve as a gold standard for measuring and guiding improvements \cite{Medsafetybench}. Evaluation metrics from synthetic question studies on TREC LiveQA and MedRedQA further reveal gaps between automated scores and human judgments, underscoring the need for nuanced, human-informed assessments \cite{safetymt1}. These evaluation insights directly inform alignment interventions. Supervised fine-tuning (SFT) uses flagged unsafe examples to reduce harmful outputs without compromising clinical accuracy, while Reinforcement Learning from Human Feedback (RLHF) treats harmful-response rates as reward signals, aiming to minimize dangerous outputs without sacrificing helpfulness. Real-time safety filters, trained on categories identified by benchmarks, add an additional safeguard by blocking risky content before delivery. Comparative research demonstrates that evaluation-driven alignment yields state-of-the-art safety in complex tasks. Direct Preference Optimization (DPO), guided by evaluation feedback, outperforms SFT in clinical reasoning, summarization, and triage \cite{safetymt2}. Advanced multi-stage pipelines—combining models such as LLaMA-2 or Mistral with preference-based fine-tuning methods 
—achieve superior safety and reliability in medical QA \cite{safetymt3}. Future work should continue leveraging evaluation-driven alignment to refine communication styles that support psychological stability in mental health contexts 
\cite{Alignment,safetymental}.

\textbf{Enhancing Explainability}
If evaluations show that a model’s answers are correct but users find them unsatisfactory due to lack of rationale, developers can incorporate techniques to force or improve explanations. One popular method is Chain-of-Thought prompting, where the model is prompted to produce step-by-step reasoning before giving the final answer. This often yields more explainable answers and can even improve accuracy. \citet{zhang2023automatic} introduces ``Let’s think step by step'' approach specifically to improve medical reasoning, which evaluation shows reduced incorrect answers and makes reasoning transparent. Another strategy is building hybrid models: e.g., first have a smaller model generate an explanation outline or causal graph, then have the main model fill in the details (as explored by \citet{Reasonmt4} with causal graphs for reasoning). \citet{hallumt3} took a different approach with interactive self-reflection: the model generates an answer, then evaluates its own answer and tries to correct any flaws, effectively explaining and refining iteratively. This showed promise in reducing reasoning errors. All these techniques are driven by recognition (through evaluation) that explainability correlates with better model understanding~\cite{MedExpQA}. Once deployed, improved explainability provides feed back: users (doctors, patients) can better identify mistakes if reasoning is visible, providing more targeted feedback for future model training.

\textbf{Improving Calibration}
Effective calibration of medical QA models begins with rigorous evaluation to identify overconfidence. Studies such as \citet{calibrmt1} have shown that across multiple specialties, current LLMs frequently assign high confidence to incorrect answers, revealing poor calibration in clinical settings. 
Benchmarks, such as MetaMedQA, further quantify these shortcomings by measuring metrics such as Confidence Accuracy and Unknown Recall, which gauge a model’s ability to recognize when it does not know the answer \cite{calibrmt3}. Similarly, QA-level calibration frameworks extend conventional reliability diagrams to entire question–answer groupings, offering theoretical guarantees that underlie more robust confidence estimates \cite{calibrmt2}. Domain-specific analyses in gastroenterology underscore these gaps: prompt-engineering and statistical methods applied to board-style questions find that even state-of-the-art LLMs struggle to represent uncertainty in a clinically meaningful way \cite{calibrmt4}. Inspired by these evaluation insights, developers employ a range of calibration techniques. Post-hoc temperature scaling or dedicated calibration training on held-out validation sets can directly reduce ECE, realigning confidence outputs with true accuracy. In generative settings, adjusting decoding parameters—such as lowering the sampling temperature—discourages the model from making overly assertive statements. Explicit prompting strategies further nudge models toward more cautious language. Beyond these, ensemble approaches and auxiliary confidence predictors offer dynamic uncertainty estimates: by aggregating outputs from multiple model instances or training a secondary classifier on question-answer pairs, the system can decide at inference time whether to hedge or assert.  Future research is poised to integrate calibration more tightly with hallucination detection—for example, by embedding two-phase verification pipelines that combine prompt engineering, statistical scoring, and consistency checks—to deliver reliable, trust-worthy medical advice under uncertainty \cite{calibrmt5}.

\section{Benchmarks and Tools for Trustworthy Medical QA}
Multiple benchmarks and evaluation tools have been developed to assess medical QA systems on the above dimensions of trustworthiness. Table~\ref{bench} provide a comparison of notable benchmarks, outlining their domain focus, format, and trustworthiness aspects they emphasize. We then highlight a few frameworks and tools that aid evaluation.

\textbf{Common Evaluation Metrics} Across these benchmarks, traditional metrics such as accuracy and precision/recall are standard for factual correctness. ROUGE/BLEU are used for comparing generated text with reference comparison, but their limitations are acknowledged~\cite{MedExQA}. To capture trust facets, some benchmarks incorporate custom metrics: e.g., Med-HALT’s false confidence rate~\cite{Med-HALT}, or MedSafetyBench’s safety score~\cite{Medsafetybench}. Human evaluation remains crucial in many benchmarks – MultiMedQA’s 12-axis rubric is administered by clinicians to rate each answer qualitatively~\cite{MedPaLM2}, and MedExQA involves human scoring of explanation correctness~\cite{MedExQA}.

\textbf{Tools and Frameworks} Beyond datasets, there are emerging tools to facilitate trustworthiness evaluation. For example, the TrustLLM Benchmark is an integrated toolkit that aggregates over 18 evaluation categories for LLMs, including medical QA scenarios ~\cite{3}. It provides a unified pipeline to test a model on many trust dimensions and compare results. Another is Holistic Evaluation of Language Models (HELM)~\cite{liang2023holistic} – not specific to medicine but often used as a template – which emphasizes transparent reporting of a model’s strengths and failures across scenarios. For explainability, some tools allow automated reasoning verification, such as checking chain-of-thought logic or using another LLM to critique the answer’s reasoning.

\section{Challenges and Future Directions} 
Despite advances in evaluation methods and benchmarks, several critical challenges remain for scalable, comprehensive assessment of medical QA systems. First, many dimensions of trustworthiness—such as clinical appropriateness, fairness, and the usefulness of explanations—still rely heavily on human expert judgment~\cite{Lekadire081554}. 
~\citet{safetymt1} and ~\citet{chowdhury-etal-2023-large} show that human evaluations often reveal subtle safety and ethics issues missed by automated tests, underscoring the necessity of expert review to ensure high-quality critique. However, 
it cannot scale to the volume of queries real systems face, and inter-rater consistency varies. 
Future work should explore automated or semi-automated proxies, for example, calibrated LLM critiques or lightweight classifiers identifying safety and bias issues. These proxies must be rigorously validated against expert evaluations to ensure reliability.

Second, existing benchmarks 
cover only a narrow set of clinical scenarios, specialties, or languages, leaving large blind spots. 
Expanding benchmark coverage through the development of multilingual medical NLP corpora is therefore a critical future direction, which is necessary to ensure equitable access to clinical AI across diverse linguistic and regional contexts. A model fine-tuned to excel on a fixed benchmark may still fail when faced with rare diseases, non-English patient queries, or emerging medical knowledge. To broaden coverage, we need dynamic, evolving datasets that incorporate real user questions
, span underrepresented specialties, and update as medical guidelines change. Projects like MedExQA, which added speech pathology, demonstrate the value of domain expansion—but many fields 
remain untested. Building flexible pipelines for continuous data collection and curation will be key.

Third, most evaluations treat each trustworthiness dimension in isolation—safety in one test, factual accuracy in another—even though these properties interact in practice. A system that maximizes safety by refusing all borderline queries may sacrifice robustness, while one that prioritizes detail could harm explainability or safety. We lack frameworks to jointly evaluate these trade-offs or to report composite trustworthiness metrics. Designing multi-objective evaluation suites—perhaps weighted ``trustworthiness scores'' co-designed with clinicians and patients—could help balance competing goals. Determining appropriate weights, however, will require careful stakeholder engagement and context-specific tailoring.
Navigating these trade-offs is highly context-dependent, as the optimal balance across dimensions varies with the application's goals and user base. For instance, a consumer-facing health tool may prioritize safety and conservative calibration, even at the cost of robustness to varied input styles. In contrast, a clinical decision support system used by experts may emphasize factuality and comprehensive coverage, operating under the assumption of expert supervision. Practitioners can also employ practical strategies to manage these trade-offs, such as applying confidence thresholds to suppress uncertain outputs, integrating retrieval-based fallback mechanisms for high-risk queries, or escalating ambiguous cases to human reviewers.

Finally, a substantial gap remains between static benchmark evaluations and real-world deployment. 
For instance, ~\citet{Towardsexpert-level} and ~\citet{1} report that despite strong benchmark performance, models like Med-PaLM exhibited overconfidence, missing context, or poor refusal behaviour in live settings with consumer health queries.
In practice, medical QA involves multi-turn conversations, clarifications, follow-up questions, and changing clinical context, dynamics rarely captured by current evaluations. 
Moreover, the real impact of errors varies widely, from harmless inaccuracies to severe consequences. 
Future research should simulate end-to-end clinical workflows—evaluating outcomes such as diagnostic accuracy, clinician efficiency, and patient satisfaction. 
Incorporating continuous user feedback loops would further align system evaluation and training with real-world needs.

\section{Conclusion}
Evaluating trustworthiness in medical QA systems involves multiple dimensions, including factuality, 
robustness, fairness, safety, explainability, and calibration. This survey reviews methods to assess each dimension and highlights current benchmarks. A key insight is that evaluation is not only measures performance but also provides critical feedback to drive improvements. We discuss examples where evaluation directly led to system enhancement. Incorporating evaluation in the development loop accelerates progress toward trustworthy QA systems suitable for critical medical use. However, current evaluations remain limited; many essential qualities are difficult to quantify, and existing benchmarks inadequately capture real-world complexity. There is substantial ongoing work needed to create more holistic and realistic evaluation frameworks, to keep pace with evolving models. 

\section*{Limitations}
In this study, we focus exclusively on medical QA systems and base our analysis on publications from 2020 and 2025. Our search covered major venues in natural language processing (e.g., ACL, EMNLP, NAACL), general AI and machine learning (e.g., NeurIPS, ICLR, AAAI), and medical informatics (e.g., JAMA, Nature Medicine, NPJ Digital Medicine). To capture cutting-edge work, we also incorporated influential preprints from arXiv. To maintain this scope, we deliberately excluded publications on general-domain LLMs and healthcare literature not directly applicable to medical QA tasks. Finally, this review is predominantly limited to English, a reflection of the current landscape, where the vast majority of benchmark datasets and clinical corpora are available in English.


\section*{Ethics Statement}
We do not see any ethics issues in this paper.

\section*{Acknowledgements}
This work was funded by 
Shandong Provincial Natural Science Foundation (project ZR2025QC636), the National Natural Science Foundation of China (projects 62372275 and 62472261), the Technology Innovation Guidance Program of Shandong Province (project YDZX2024088), and the Provincial Key R\&D Program of Shandong Province (project 2024CXGC010108). The research was also partially funded by The Natural Sciences and Engineering Research Council of Canada (NSERC) through a Discovery Grant to R. E. Mercer, and F. Rudzicz is supported by a CIFAR Chair in AI.

\bibliography{custom}

\appendix

\section{Appendix}

\label{sec:appendix}

\begin{table*}
  \centering
  \scalebox{0.89}{
  {\fontsize{9pt}{9pt}\selectfont
  \begin{tabular}{>{\centering\arraybackslash}m{2.5cm} >
   {\centering\arraybackslash}m{4.5cm}>
   {\centering\arraybackslash}m{5.8cm} >
   {\centering\arraybackslash}m{3cm}}
    \hline
    \specialrule{0.8pt}{0pt}{1pt}
    \textbf{{\large Benchmark}}   & \textbf{{\large Description}} & \textbf{ {\large Key trustworthiness focus} } &\textbf{{\large Format }} \\
    \hline
    \specialrule{0.8pt}{0pt}{1pt}
    {\normalsize Med-HALT } &   Medical Hallucination Test dataset. Derived from medical exams across countries to probe factual recall and reasoning.   &    \textbf{Factuality:} includes reasoning-based tests (“False Confidence”, ”None of the above“trick questions) and memory-based recall tests to quantify hallucination rates. Evaluates how often models produce unsupported info under stress-test conditions.   &   Multiple-Choice Questions, Yes/No, Open-ended question   \\     \hline
    
   {\normalsize MedHallBench } & A comprehensive medical hallucination evaluation framework integrating automated clinical medical image caption hallucination scoring (ACHMI) and clinical expert review.    &  \textbf{Factuality:} Design questions centered around object hallucinations, attribute hallucinations, multimodal conflict hallucinations, and logical reasoning hallucinations, and conduct adversarial tests to uncover the causes of hallucinations in models.   &   Open-ended Q\&A, Visual Question Answering, Summarization     \\        
    \hline
    
    {\normalsize MedHallu}  &  The first binary classification benchmark for medical hallucination detection. The questions are divided into three levels - Easy, Medium, and Hard - according to the difficulty of identifying hallucinations.    &  \textbf{Factuality:} Detect whether the model can correctly classify the labels of question-answer pairs as "real" or "hallucination".   &    Binary Hallucination Detection    \\        
    \hline
    
    \specialrule{0.6pt}{0pt}{1pt}
    {\normalsize MedFuzz}  &  By applying adversarial perturbations to medical question-answering queries, evaluate the robustness and performance of large language models (LLMs) in medical question-answering tasks.   &  \textbf{Robustness:} In the evaluation, first input the correct questions and answers into the model. Then, use the Attacker LLM to modify the original questions for multiple rounds and input them into the model. Each modification attempts to guide the target model to select the wrong answer without changing the correct answer of the original question.   &    Multiple-Choice Questions    \\        
    \hline
    \specialrule{0.6pt}{0pt}{1pt}
    { \normalsize BiasMedQA }  &  {\fontsize{8.5pt}{9pt}\selectfont A benchmark dataset for evaluating whether there is bias (towards different patient groups such as those of different genders, races, etc.) in LLMs in medical question answering. }  &   \textbf{Fairness:} Introduce common clinically relevant cognitive biases into USMLE questions to test the performance of the model when facing these biases.  &   Multiple-Choice Questions     \\        
    \hline
    \specialrule{0.6pt}{0pt}{1pt}
   { \normalsize MedSafetyBench }  &   The first medical-domain Safety evaluation benchmark dataset focused on assessing model responses to unsafe medical instructions.   &  \textbf{ Safety:} Evaluate whether models can ensure response integrity when handling inputs containing unsafe medical instructions, as benchmarked by MedSafetyBench's adversarial testing framework.   &    Open-ended Q\&A    \\        
    \hline
    \specialrule{0.6pt}{0pt}{1pt}
    { \normalsize MedExQA } &  Medical explainability QA benchmark. Covers 5 underrepresented specialties (e.g. speech pathology, clinical psych) with multiple ground-truth explanations per Q\&A.   &  \textbf{ Explainability: } evaluates if models can provide nuanced medical explanations beyond just correct answers. Uses lexical metrics and human ratings to score explanation quality. Also tests knowledge in less-studied specialties (robustness to specialty domains).   &  Open-ended question, required free-text explanation for answer.    \\        
    \hline
     { \normalsize PubMedQA } &  A Medical Reasoning Evaluation Benchmark for LLMs that Combine Expert-Annotated and Automated Knowledge Expansion, designed to assess contextual reasoning capabilities across medical texts and domain knowledge.      &   \textbf{ Explainability:} Given a question and a medical text context with the conclusion section removed, evaluate whether the model can infer if the question originally appeared in the conclusion section of the source text.  &  Three-way classification \\        
    \hline
    { \normalsize DR.BENCH } &   A benchmark for evaluating clinical diagnostic reasoning capabilities of large language models (LLMs), comprising six reasoning tasks: MedNLI, Assessment and Plan Relation Labeling, EmrQA, SOAP Section Classification, Problem Summarization, and Diagnosis Generation.    &  {\fontsize{8.3pt}{8.5pt}\selectfont \textbf{ Explainability:} The six diagnostic reasoning task categories in DR.BENCH comprehensively span the clinical workflow-continuum, designed to evaluate the model's capabilities including: medical concept logic; context-aware information retrieval; structured clinical knowledge classification; knowledge-graph-driven causal reasoning; multi-step evidence integration; knowledge-intensive clinical inference. }  &    Multiple-Choice Questions, Extractive QA, Open-ended Questions, Text Generation    \\        
    \hline
     { \normalsize MedExpQA }  &  MedExpQA encompasses multiple languages. For each question, a standard answer is provided along with multiple Gold-Explanation explanations written by medical experts.   & {\fontsize{8.3pt}{8.5pt}\selectfont \textbf{ Explainability:} Three types of tasks are set during evaluation: basic input only, basic input plus gold-standard explanation, and basic input plus RAG text. By comparing the outputs of the three types of tasks, the amount of missing reasoning ability of the model and the degree of help of automatically retrieved knowledge for the model's reasoning can be evaluated. }  &   Multiple-Choice Questions  \\   
    \hline
    { \normalsize MediQ } &  A benchmark evaluating LLMs' capabilities in reliable interactive clinical reasoning, designed to assess their reasoning abilities by observing performance on informationally incomplete clinical queries.   &  \textbf{ Explainability:} Evaluating the simulation of a dynamic clinical interaction environment where the model under assessment acts as an Expert System, with performance under informationally incomplete initial conditions recorded to measure interactive clinical reasoning capabilities.  &   Multiple-Choice Questions, Interactive Q\&A  \\    
    \hline
    
   { \normalsize MedXpertQA } &   A comprehensive benchmark for assessing expert-level medical knowledge and advanced reasoning capabilities, comprising Text (text-based) and MM (multimodal) subsets, with an independently designed reasoning subset.   &  \textbf{Explainability:} The reasoning subset comprises highest-difficulty questions requiring multi-step logical reasoning, selected from both Text (text-based) and MM (multimodal) configurations, specifically designed to evaluate model reasoning capabilities   &   Multiple-Choice Questions, Multimodal QA     \\           
    \hline
    \specialrule{0.8pt}{0pt}{0pt}

  \end{tabular}
  }
  }
  \caption{\label{bench}
    Summary of representative benchmarks for each dimension, including their descriptions, key trustworthiness focus, and data format.
  }

\end{table*}

\end{document}